\documentclass[runningheads]{llncs}

\usepackage[T1]{fontenc}

\usepackage{amsmath}
\usepackage{booktabs}
\usepackage{hyperref}
\usepackage{cleveref}
\usepackage{delimset}
\usepackage{glossaries}
\usepackage{graphicx}
\usepackage{verbatim}
\usepackage{xcolor}
\usepackage{makecell}
\usepackage{tabularx}
\usepackage{color}

\begin{document}
\title{Multi-Dataset Diagnostic Utility of Clinical Visual Concepts in AI Systems for Dermatology}

\newacronym{ai}{AI}{artificial intelligence}
\newacronym{ci}{CI}{confidence interval}
\newacronym{ap}{AP}{average precision}
\newacronym{llm}{LLM}{large language model}
\newacronym{pca}{PCA}{principal component analysis}
\newacronym{rocauc}{ROC-AUC}{receiver operating characteristic area under the curve}
\newacronym{vqa}{VQA}{visual question answering}

\author{
Linda Wermelinger\inst{1,2} \and
Simone Lionetti\inst{2} \and
Fabian Gr\"oger\inst{1,2} \and
Nipun Ranasekara\inst{1,2}\and
Philippe~Gottfrois\inst{1,3} \and
Ludovic Amruthalingam\inst{2} \and
Labelling~Consortium\inst{3,\dagger} \and
Marc Pouly\inst{2} \and
Alexander A. Navarini\inst{1,3}
}

\authorrunning{L. Wermelinger et al.}
\titlerunning{Multi-Dataset Diagnostic Utility of Clinical Visual Concepts}

\institute{University of Basel \and
Lucerne University of Applied Sciences and Arts \and
University Hospital of Basel \\[0.3em]
\textdagger\ Thierry Grimm, Bianca Wahrenberger, Maximilian Kass \\[0.3em]
\email{linda.wermelinger@hslu.ch}}
  
\maketitle              
\begin{abstract}

The clinical integration of AI systems in digital dermatology relies heavily on human trust.
Clinically interpretable visual concepts can act as intermediate representations enhancing trust and reliability.
However, research in this domain is currently limited by scattered, heterogeneous dataset annotations.
In this work, we introduce SkinLex, a harmonized dataset of 48 clinical morphological attributes across four public datasets (SkinCon, DermaCon-IN, MM-Skin, and PASSION) for a total of 20,411 records.
Supervised nine-partition classification of skin conditions shows that limiting features to specific visual groups, like shapes or colors alone, reduces diagnostic accuracy.
Bootstrapped backward elimination reveals that the set of 48 visual concepts has some degree of redundancy for algorithmic nine-partition diagnosis on the examined dataset.
This demonstrates that coarse diagnosis on the selected dataset requires a relatively small but varied combination of clinical concepts, and motivates further research to improve concept taxonomy.
Results can be translated into clinical benefits by reducing inputs for concept-based models, improving efficiency for annotation and modeling, and further enhancing interpretability.
Code and prompt templates are available at \url{https://github.com/Digital-Dermatology/SkinLex}.


\end{abstract}

\section{Introduction}

\begin{figure}[t]
    \centering
    \includegraphics[width=\textwidth]{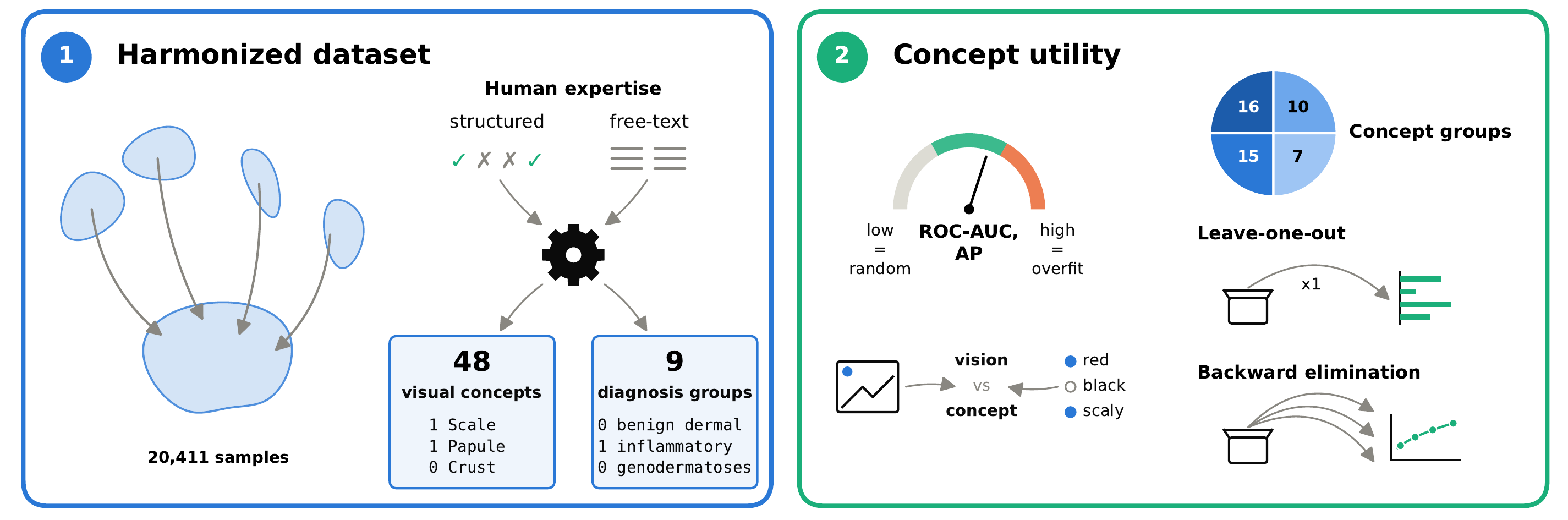}
    \caption{
    Overview of the paper: the harmonized SkinLex dataset and visual concept utility analysis.
    }
    \label{fig:abstract}
\end{figure}

The clinical adoption of \gls{ai} decision-support systems in digital dermatology depends heavily on human trust \cite{chanda_dermatologist-like_2024}. 
Clinical concepts are emerging as an intermediate layer that can corroborate trust and steer decisions or corrections, as they enable intuitive audits by medical staff \cite{koh_concept_2020,pang_integrating_2024}.
One prominent approach in this direction is concept bottleneck modeling, which uses human-interpretable concepts as an interpretable layer \cite{koh_concept_2020}.
Recent work extends this idea by integrating clinical knowledge into concept bottleneck models \cite{pang_integrating_2024} and refining concept-based representations for medical image diagnosis \cite{wang_concept_2026}.

Dermatological examinations routinely use a standardized vocabulary to describe skin disorders by assessing primary lesions, their secondary changes, and their morphology (color, size, shape, surface), distribution, and progression \cite{lipsker_clinical_2013}.
In digital dermatology, this was first incorporated in dermatoscopic settings through early specialized resources like PH$^2$ \cite{mendonca_ph_2013} and Derm7pt \cite{kawahara_7-point_2018}.
Both datasets adopt specific, expert-annotated dermatoscopic structures, such as global asymmetry or the 7-point checklist criteria \cite{argenziano_epiluminescence_1998,mendonca_ph_2013,kawahara_7-point_2018}.

Subsequent work expanded this to broader clinical dermatology through datasets with structured concept metadata.
For instance, SkinCon \cite{daneshjou_skincon_2022} provides expert annotations for 48 morphological features across a subset of images from Fitzpatrick17k \cite{groh_evaluating_2021} and DDI \cite{daneshjou_disparities_2022}.
These descriptors consist of clinical terms that capture primary lesions, secondary changes, shapes, and colors.
This concept set is closely mirrored by DermaCon-IN \cite{madarkar2026dermaconin} across a patient cohort from outpatient clinics in South India, adding features like striae and nail abnormalities.
While both SkinCon and DermaCon-IN rely on manual, expert-validated annotations, other approaches like MILK10k \cite{tschandl_milk10k_2026} automate concept probability extraction using an image-text foundation model trained on medical literature \cite{kim_transparent_2024}.

Driven by the success of vision-language models, recent work has increasingly turned toward datasets featuring detailed, natural language clinical descriptions rather than rigid categorical labels.
While not concept datasets in the strict sense, their captions contain recoverable morphological information that maps to a shared concept vocabulary. 
Examples include MM-Skin \cite{zeng_mm-skin_2025}, a dataset pairing diverse modalities with long-form clinical captions, and Derm1M \cite{yan_derm1m_2025}, which uses a 130-item list to extract visual characteristics from raw text.
Similarly, the PASSION dataset \cite{Got_PASSION_MICCAI2024} of sub-Saharan skin conditions was recently extended with dual expert captions \cite{voegtli2026passion}.

While several public datasets now provide conceptual metadata, progress is hindered by the scattered and heterogeneous character of this concept-related information.
Such heterogeneity is understandable, as human annotations require substantial expert effort to collect \cite{daneshjou_skincon_2022,zhou_skincap_2024}.
Moreover, dermatology datasets are collected across diverse patient populations and clinical settings, often reflecting locally defined label sets and annotation practices \cite{groger_global_2026,daneshjou_disparities_2022}.
As a result, manually annotated concept information is available only for a relatively small fraction of digital dermatology images in formats that have seen little standardization and are typically sparse, complicating large-scale analyses.
Such limitations create a need for cross-dataset concept harmonization before questions about concept redundancy, complementarity, and predictive utility can be studied at scale.

So far, the data-driven methods community has leveraged structured concepts from clinical dermatology to develop interpretable, fair and auditable \gls{ai} systems \cite{pang_integrating_2024,daneshjou_skincon_2022,daneshjou_checklist_2022}.
In this work, we simultaneously increase the scale of this endeavor and take first steps toward exploring how these models can provide valuable, interpretable feedback to dermatological practice.
By harmonizing concept annotations across four dataset sources, we open the path to a broader analysis of clinical attributes in digital dermatology.
At the same time, we train, evaluate, and examine how machine-learning models predict skin diseases directly from these structured concepts.
In doing so, we explore the assumption that a unified set of clinical concepts can provide a shared feature representation for computer-assisted diagnosis across diverse disease categories, treating concepts as a primary input modality rather than mere explanatory support.
Our empirical study of model behavior informs concept-bottleneck approaches in first instance, but our hope is that, with additional evidence from future work, it could guide how visual concepts are selected and annotated in digital dermatology.

The main contributions of this paper, illustrated in \cref{fig:abstract}, are as follows.
\begin{enumerate}
\item We introduce SkinLex, a unified dataset built from sources that rely on human expertise (\cref{ssec:harmonization}), which feature either structured annotations by clinicians (SkinCon, DermaCon-IN) or expert-authored text (MM-Skin, PASSION).
We map clinical visual concepts to the set established by SkinCon, using exact matching where possible or processing free text with a \gls{llm}.
We consolidate more than 6,000 noisy, fine-grained diagnoses into the nine Fitzpatrick17k categories leveraging the same \gls{llm}.
This produces a dataset with structured visual clinical concepts and diagnostic information derived from human expertise that is five times larger than SkinCon.
\item We evaluate the utility of the 48 selected concepts for predicting the nine diagnostic categories across the unified dataset.
We establish metrics including lower and upper bounds (\cref{ssec:setup}), and compare concept performance to deep visual feature extraction (\cref{ssec:reference}).
We then evaluate different groups of concepts such as shapes or colors, assess how critical individual concepts are (\cref{ssec:groups}), and estimate the degree of redundancy in the concept set (\cref{ssec:elimination}).
Results indicate that classification requires varied concept categories to work well and that some redundancy is present.
\end{enumerate}
Together, these two contributions facilitate the evaluation of clinical visual concepts for computer-assisted diagnosis by separating concept utility from their image-based extraction, and trace a path for the refinement of concept sets.

\section{Methodology}

\subsection{Dataset Harmonization}
\label{ssec:harmonization}

\begin{table}[t]
\centering
\footnotesize
\caption{Summary of original and final sample counts across unified sources.}
\label{tab:dataset_overview}
\setlength{\tabcolsep}{3pt}
\renewcommand{\arraystretch}{1.2}
\begin{tabularx}{\linewidth}{
  l 
  >{\hsize=1.1\hsize\raggedright\arraybackslash}X
  c
  c
  >{\hsize=0.9\hsize\raggedright\arraybackslash}X
}
\toprule
\textbf{Source Dataset} & \textbf{Image Modality} & \textbf{Original} & \textbf{Final} & \textbf{Alignment} \\
\midrule
SkinCon & Clinical & \phantom{0}3,886 & 3,642 & Direct match. \\
DermaCon-IN & Clinical & \phantom{0}5,450 & 5,405 & Direct match. \\
MM-Skin & Clinical, Dermatoscopy, Pathology & 11,015 & 7,500 & Concept and disease extraction.\\
PASSION & Clinical & \phantom{0}5,127 & 3,864 & Concept extraction.\\
\midrule
\textbf{SkinLex} & \textbf{Mixed} & \textbf{25,478} & \textbf{20,411} & \\
\bottomrule
\end{tabularx}
\end{table}

Previous works have investigated the capabilities and limitations of concept-based methods, but these analyses are either performed on small datasets or entangled with the ability of complex models to correctly infer such concepts from images.
To obtain a strong data signal and circumvent image-based concept inference, we aggregate and harmonize information from several sources.
In doing so, we strive to obtain sufficiently clean diagnostic information that can be used for a unified analysis.
However, direct human annotations for concepts and diagnoses are sparse, so we leverage the human signal from image captions that describe skin conditions when structured labels are not available.
To this end, we use \texttt{QuantTrio/Qwen3.6-35B-A3B-AWQ}~\cite{qwen36_35b_a3b} to extract concepts and diagnoses from text, where they can often be found without requiring complex processing.
We use zero temperature to minimize sampling variability, refer to this model simply as ``Qwen'' in the following, and provide all prompt templates in the project's GitHub repository.

To construct SkinLex with a unified visual clinical vocabulary, we map each sample onto the SkinCon concept space (e.g., erythema, scale, plaque) which is based on dermatology textbook definitions~\cite{Bolognia_Schaffer_Cerroni_2017}.
Two of the sources already provide concepts as binary 0/1 labels in this space, while for others we extract the 48 SkinCon concepts in the same format using Qwen.
Direct alignment of the raw diagnostic labels is impractical because of different granularity, terminology, and frequency across sources, which results in more than 6,000 separate condition identifiers.
We therefore map these identifiers to a common category set using Qwen, generating a dictionary from the diagnostic labels of each dataset when possible, and otherwise leveraging human-provided captions which often directly include information on the condition.
Given the size of the final dataset, we settle for the nine-partition mutually-exclusive classification that was established with Fitzpatrick17k.
To ensure medical grounding and reduce the risk of hallucination, we include the text-based granular-to-nine-partition mapping dictionary from the Fitzpatrick17k dataset as in-context reference.

The details of sources, harmonization, and quality checks are as follows.
\begin{itemize}
    \item \textbf{SkinCon}: Samples are filtered using the CleanPatrick list \cite{groger_cleanpatrick_2026} and the label ``Do not consider this image'', leaving 3,642 instances.
    \item \textbf{DermaCon-IN}: Images with insufficient quality for diagnosis according to the metadata and entries labeled ``No Definite Diagnosis'' are removed, leaving 5,405 instances.
    \item \textbf{MM-Skin}: Discrete clinical concepts and granular diagnostic labels are extracted from captions and \gls{vqa} dialogues using Qwen. Although MM-Skin includes dermatoscopic and pathological images alongside clinical photographs, captions across these modalities contain overlapping concept sets. We discard extraneous files, as well as cases where the model extracted multiple or no diagnoses, leaving 7,500 instances.
    \item \textbf{PASSION}: We extract clinical concepts from the two expert captions per sample using Qwen. This dataset has structured diagnostic annotations, so we only drop instances marked as unspecified and retain 3,864.
\end{itemize}
We do not include Derm1M because 378,678 of its 423,016 captions are repeated identically at least two times, with the most frequent appearing in 2,945 copies.
This produces many entries where concept set and diagnosis are exactly the same and would lead to unrealistic reweighting with no variance.

The final sample counts and alignment methods are summarized in \cref{tab:dataset_overview}, and the diagnosis distribution according to the nine-partition Fitzpatrick17k classification is given in \cref{tab:nine_partition_distribution}.
While individual sources exhibit severe localized imbalances, their aggregation fills critical data gaps.
For instance, features completely absent ($0.0\%$) from DermaCon-IN such as \textit{Sclerosis}, \textit{Blue}, and \textit{Warty/Papillomatous} morphologies are heavily supplemented by MM-Skin (accounting for $87.1\%$, $96.7\%$, and $79.9\%$ of those concepts, respectively).
Likewise, the PASSION dataset provides $92.4\%$ of all \textit{Excoriation} instances.
This improvement in concept and diagnosis coverage provides a basis for novel analyses.

\begin{table}[t]
\centering
\footnotesize
\caption{Distribution of the aligned disease categories across the unified dataset.}
\label{tab:nine_partition_distribution}
\setlength{\tabcolsep}{8pt}
\renewcommand{\arraystretch}{1.1}
\begin{tabularx}{\linewidth}{Xrr}
\toprule
\textbf{Standardized Disease Category} & \textbf{Sample Count} & \textbf{Percentage (\%)} \\
\midrule
Inflammatory           & 15,091 & 73.94 \\
Malignant Epidermal    &  964 &  4.72 \\
Genodermatoses         &   867 &  4.25 \\
Benign Dermal          &   865 &  4.24 \\
Benign Epidermal       &   825 &  4.04 \\
Benign Melanocyte      &   641 &  3.14 \\
Malignant Cutaneous Lymphoma    &   565 &  2.77 \\
Malignant Melanoma     &   465 &  2.28 \\
Malignant Dermal       &   128 &  0.63 \\
\midrule
\textbf{Total} & \textbf{20,411} & \textbf{100.00} \\
\bottomrule
\end{tabularx}
\end{table}

\subsection{Experimental Setup}
\label{ssec:setup}

The SkinLex dataset is split randomly with stratification into training ($64\%$), validation ($16\%$), and testing ($20\%$) subsets.
Because of imbalance, with the largest diagnosis accounting for $73.94\%$ of the samples, we evaluate the models both using macro-averaged \gls{rocauc} and \gls{ap}.
These metrics bypass threshold calibration and have baselines of 0.5000 and 0.1111 respectively.
Since the concept presence space is sparsely populated and some samples with different diagnoses might not be separable, we establish a performance ceiling by overfitting the validation set, obtaining 0.9404 \gls{rocauc} and 0.5568 \gls{ap}.

We investigate how models utilize different clinical concepts for diagnosis.
First, we restrict the input feature space to one of four clinically defined feature groups: \textit{Colors} (10 features), \textit{Shapes} (16 features), \textit{Textures} (15 features), and \textit{Other} (7 features) to measure their standalone predictive capacity.
Second, we perform a leave-one-out sensitivity analysis by removing one clinical concept at a time and retraining 48 models with fixed hyperparameters.

To estimate the potential redundancy of the 48-concept set for the coarse diagnosis task under exam, we run backward elimination with 100 bootstrapped training and validation sets, each with a separate sample pool.
At each step, the feature whose individual exclusion produces the best median score across bootstraps is removed until the feature set is empty.
The last zero-feature model is a dummy classifier that randomly assigns labels with the training set frequency.

The minimal concept set that retains the full performance is determined by a one-sided paired Wilcoxon signed-rank test on the bootstrapped scores with a $p$-value threshold of 5\%.
Score differences are indeed symmetric under the null hypothesis that the model trained on the reduced concept set is not worse than the one using the full 48-concept set.
No correction is applied for multiple testing, as lowering the threshold would increase the chance of no observed difference, which could inflate the set of concepts deemed redundant.

\section{Experiments and Results}

\subsection{Reference models}
\label{ssec:reference}

To select promising models for concept-based nine-partition diagnosis, we screen twelve classic candidates on the validation set using PyCaret \cite{PyCaret}.
LightGBM~\cite{ke_lightgbm_2017} achieves the best results, capturing $90.9\%$ of the \gls{rocauc} and $58.1\%$ of the \gls{ap} ceiling with 0.8555 and 0.3236 respectively.
We therefore use LightGBM as the primary model, and include one-versus-rest logistic regression (\gls{rocauc}: 0.8384, \gls{ap}: 0.2947) as a linear comparison.
For reference, we tested the same models on classification with frozen image feature sets extracted by ViT-base~\cite{dosovitskiy_vit_2021} models.
A LightGBM classifier on DINOv3 features~\cite{simeoni_dinov3_2025} obtains 0.8907 \gls{rocauc} and 0.3756 \gls{ap}, outperforming supervised ImageNet features.

\subsection{Concept Groups and Sensitivity Analysis}
\label{ssec:groups}

\begin{table}[t]
\centering
\newcommand{\cell}[3]{$#1\,\brk[s]*{{}_{#2}^{#3}}$}
\footnotesize
\caption{Model performance across full and restricted concept groups. Metrics report median and [95\% \gls{ci}] estimated with 100 validation bootstraps.}
\label{tab:group_performance}
\begin{tabular*}{\linewidth}{l@{\extracolsep{\fill}}ccccc}
\toprule
 & & \multicolumn{2}{c}{LightGBM} & \multicolumn{2}{c}{Logistic Regression} \\
\cmidrule(r){3-4} \cmidrule(l){5-6}
Feature Set & Features & \gls{rocauc} & \gls{ap} & \gls{rocauc} & \gls{ap} \\
\midrule
Full Model  & 48  & \cell{0.829}{0.812}{0.845} & \cell{0.309}{0.288}{0.330} & \cell{0.801}{0.786}{0.819} & \cell{0.274}{0.255}{0.299} \\[4pt]
\midrule
Shapes      & 16  & \cell{0.711}{0.695}{0.727} & \cell{0.174}{0.163}{0.183} & \cell{0.672}{0.658}{0.689} & \cell{0.170}{0.158}{0.181} \\[4pt]
Colors      & 10  & \cell{0.690}{0.671}{0.713} & \cell{0.173}{0.164}{0.185} & \cell{0.683}{0.665}{0.707} & \cell{0.177}{0.165}{0.191} \\[4pt]
Textures    & 15  & \cell{0.661}{0.644}{0.675} & \cell{0.153}{0.145}{0.163} & \cell{0.659}{0.643}{0.675} & \cell{0.155}{0.148}{0.163} \\[4pt]
Others      & 7   & \cell{0.530}{0.524}{0.538} & \cell{0.124}{0.119}{0.129} & \cell{0.532}{0.526}{0.541} & \cell{0.124}{0.119}{0.129} \\
\bottomrule
\end{tabular*}
\end{table}

\Cref{tab:group_performance} shows that restricting inputs to isolated concept groups considerably degrades performance.
The \textit{Shapes} and \textit{Colors} groups preserve the strongest standalone signals, while \textit{Others} approaches random guessing.
Altogether, results highlight the need for a diverse, mixed concept set.

\begin{figure}[t]
    \centering
    \includegraphics[width=\textwidth]{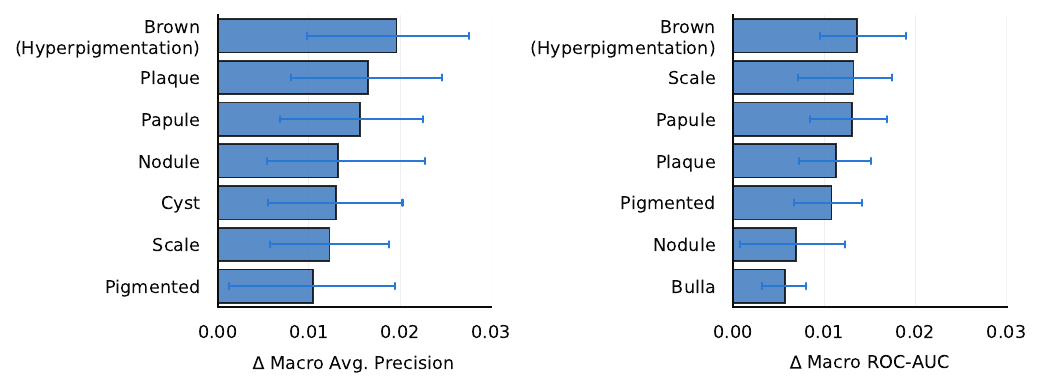}
    \caption{
    Leave-one-out sensitivity for LightGBM on the validation split.
    The bars illustrate the mean absolute decrease in macro \gls{ap} (left) and \gls{rocauc} (right) when a concept is omitted, with higher values indicating concepts critical for performance.
    Error bars denote the standard deviation across 100 validation bootstraps.
    }
    \label{fig:sensitivity_analysis}
\end{figure}

Concurrently, the leave-one-out sensitivity in \cref{fig:sensitivity_analysis} demonstrates individual concept redundancy, as performance shifts are small.
The color \textit{Brown (Hyperpigmentation)} leads to the largest degradation, followed by structural attributes like \textit{Plaque}, \textit{Papule}, \textit{Nodule}, and \textit{Scale}.

\subsection{Backward Elimination}
\label{ssec:elimination}

\begin{figure}[t]
    \centering
    \includegraphics[trim=0.3cm 0.3cm 0.3cm 0.3cm,clip,width=0.49\textwidth]{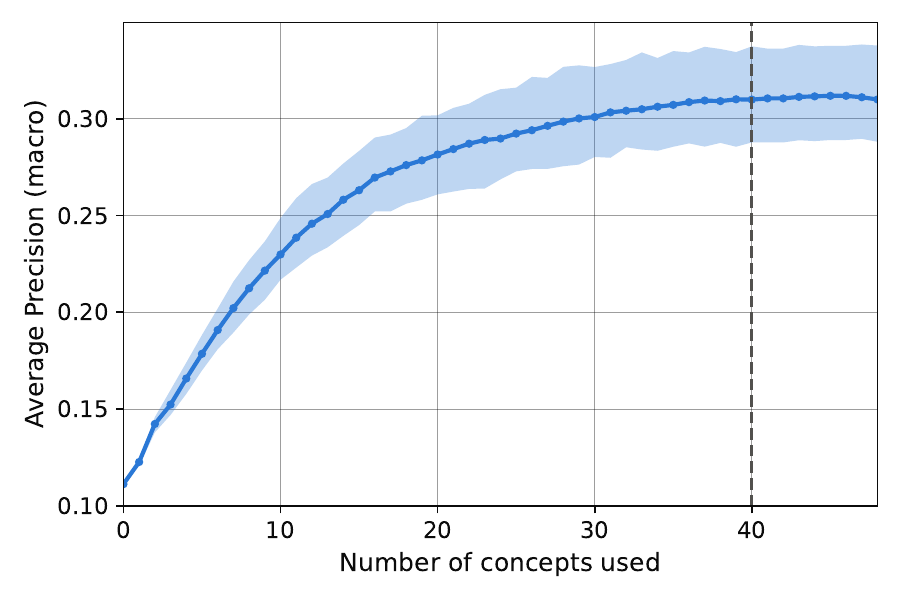}
    \includegraphics[trim=0.3cm 0.3cm 0.3cm 0.3cm,clip,width=0.49\textwidth]{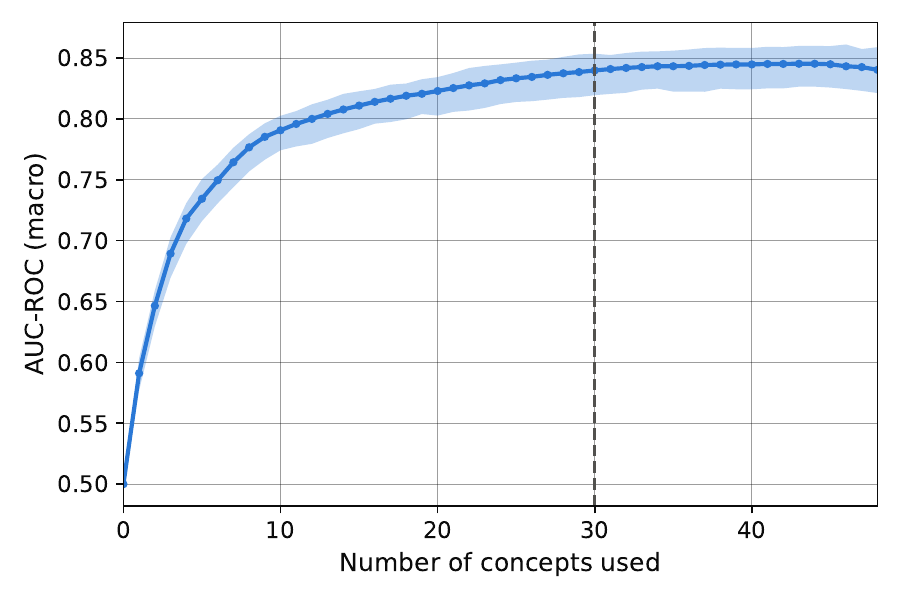}
    \caption{
    Median and 95\% \gls{ci} of LightGBM macro \acrshort{ap} and \gls{rocauc} across the number of concepts used, using 100 bootstraps of training and validation sets.
    The dashed line indicates the highest number of used concepts where a paired one-sided Wilcoxon signed-rank test indicates lower performance than the full set.
    }
    \label{fig:trajectory_analysis}
\end{figure}

\Cref{fig:trajectory_analysis} illustrates model performance in terms of macro-averaged \gls{rocauc} and \gls{ap} as a function of the number of concepts in the backward-elimination process.
As expected, the returns in terms of scores grow sub-linearly with the number of concepts used for classification.
The statistical test indicates that removing 8 concepts does not significantly lower \gls{ap} compared to the full concept set, while up to 18 concepts can be omitted without decreasing \gls{rocauc}.
These numbers suggest some redundancy for algorithmic diagnosis in this dataset, although the outcome depends on the evaluation metric.
Note that it is not possible to deduce if and which concepts lack utility from the result, as a missing concept can be compensated for by others, and which one is removed may be an artifact.

\section{Discussion}

\subsection{Limitations}

First, our flat 48-concept set and nine-diagnoses partition represent a simplified setup that may not reflect consensus across medical institutions, as it omits disease hierarchies and context-dependent clinical vocabularies.
Finer concept sets, such as those in Derm1M, or more granular disease targets may capture greater diagnostic nuance and increase task complexity.
Although the concept set focuses on visual descriptions rather than diagnostic labels, some annotation bias toward specific conditions may be present.

Second, dataset harmonization carries extraction constraints.
Because original sources emphasize concept presence over absence, caption-based extraction is prone to conflating unmentioned features with negative findings.
Additionally, automated extraction and harmonization via Qwen require further cross-model and human verification.

Finally, caution is required when generalizing these findings.
While combining diverse sources improves overall coverage, it likely introduces hidden subgroups and demographic constraints.
Humans or alternative models may leverage concepts differently, yielding distinct importance rankings.
Even within the present scope, a concept's lack of utility for algorithmic coarse-grained classification cannot be interpreted as general low relevance, as redundancy is only defined within a set of concepts.

\subsection{Conclusions and Outlook}

In this work, we introduce SkinLex, a dataset harmonizing four skin image datasets with human annotations of visual concepts and diagnostic labels.
By merging these into a single collection of 20,411 samples with structured information, we enabled an analysis of concept utility in a specific diagnostic task.
Empirical results confirm that different concept categories such as shape and color supply complementary information, and suggest that the 48 concepts reported by SkinCon have some level of redundancy for coarse algorithmic classification.

These findings carry clinical relevance for explainable AI deployment.
Identifying and removing redundant concepts can reduce annotation effort, improve computational efficiency, elevate interpretability and increase trust in clinical environments by presenting a concise, stable core of diagnostic indicators rather than a noisy list of low-signal terms.

\medskip
\noindent\textbf{Acknowledgments.}
This research was funded by the Swiss National Science Foundation (SNSF) under grant 20HW-1 228541.

\smallskip
\noindent\textbf{Disclosure of Interests.}
A.A.N. is a shareholder of Derma2Go AG and Derma\allowbreak.One GmbH. All other authors declare no competing interests.

%
%
%
\bibliographystyle{splncs04}
\bibliography{bibliography.bib}

\end{document}